\pdfoutput=1

\documentclass[11pt]{article}

\usepackage[preprint]{acl}

\usepackage{times}
\usepackage{latexsym}
\usepackage{hyperref}

\usepackage[T1]{fontenc}

\usepackage[utf8]{inputenc}

\usepackage{microtype}

\usepackage{inconsolata}

\usepackage{graphicx}
\usepackage{amsmath}

\usepackage{hyperref}
\usepackage{url}

\usepackage{graphicx}
\usepackage{wrapfig}
\usepackage{tikz}
\usepackage{subfigure}
\usepackage{float}
\usepackage{caption}
\usepackage{enumerate}
\usepackage{array}
\usepackage[ruled, linesnumbered]{algorithm2e}
\usepackage{hyperref}
\usepackage{url}
\usepackage{todonotes}
\usepackage{ulem}

\usepackage{colortbl}

\usepackage{rotating}
\usepackage{booktabs}
\usepackage{multirow}
\usepackage{makecell}
\usepackage{amssymb}

\usepackage{amsmath}

\usepackage{amssymb}
\usepackage{pdfrender}

\usepackage{xcolor}

\SetKwInput{KwInput}{Input}                
\SetKwInput{KwOutput}{Output}              
\usepackage{algorithmic}

\title{Do Large Language Models Possess Sensitive to Sentiment?}

\author{Yang Liu, Xichou Zhu, Zhou Shen, Yi Liu, Min Li, Yujun Chen,
Benzi John, \\
\textbf{Zhenzhen Ma, Tao Hu, Zhi Li, Zhiyang Xu, Wei Luo, Junhui Wang}\\
Machine Learning \& AI Team, Privacy and Data Protection Office\\
  ByteDance, Beijing, China\\
  \texttt{\{liuyang.173, junhui.wang\}@bytedance.com} 
}

\begin{document}
\maketitle
\begin{abstract}

Large Language Models (LLMs) have recently displayed their extraordinary capabilities in language understanding. However, how to comprehensively assess the sentiment capabilities of LLMs continues to be a challenge. This paper investigates the ability of LLMs to detect and react to sentiment in text modal. As the integration of LLMs into diverse applications is on the rise, it becomes highly critical to comprehend their sensitivity to emotional tone, as it can influence the user experience and the efficacy of sentiment-driven tasks. We conduct a series of experiments to evaluate the performance of several prominent LLMs in identifying and responding appropriately to sentiments like positive, negative, and neutral emotions. The models' outputs are analyzed across various sentiment benchmarks, and their responses are compared with human evaluations. Our discoveries indicate that although LLMs show a basic sensitivity to sentiment, there are substantial variations in their accuracy and consistency, emphasizing the requirement for further enhancements in their training processes to better capture subtle emotional cues. Take an example in our findings, in some cases, the models might wrongly classify a strongly positive sentiment as neutral, or fail to recognize sarcasm or irony in the text. Such misclassifications highlight the complexity of sentiment analysis and the areas where the models need to be refined. Another aspect is that different LLMs might perform differently on the same set of data, depending on their architecture and training datasets. This variance calls for a more in-depth study of the factors that contribute to the performance differences and how they can be optimized. 

\end{abstract}

\section{Introduction}

Recently, large language models (LLMs) have made groundbreaking strides that have dramatically reshaped the artificial intelligence landscape \citep{brown2020language, chowdhery2023palm}. These models have become a cornerstone in natural language processing (NLP), enabling advances in various tasks, from text generation \citep{mo2024large, li2024dtllm, abburi2023generative} to question answering \citep{zhuang2024toolqa, saito2024unsupervised}. Despite their widespread adoption, one crucial area that remains insufficiently explored is their capability to accurately perceive and respond to sentiment. Sentiment analysis—the process of identifying the emotional tone within text—is vital for applications such as customer feedback analysis, social media monitoring, and conversational agents \citep{liao2023proactive, zhang2024s}. This raises an important question:


\noindent\fbox{ \parbox{0.95\columnwidth}{
\texttt{
Are these advanced LLMs, trained on extensive datasets, truly capable of understanding sentiment, or are they simply replicating the sentiment patterns they have learned from their training data?
}
}}
\\
 

This paper aims to thoroughly evaluate the performance of large language models (LLMs) in sentiment analysis, specifically assessing their ability to detect and generate responses that correspond to the sentiment present in the input text. We explore models with varying architectures and sizes to identify both the strengths and the areas needing improvement in sentiment-related tasks. Our evaluation process follows a structured workflow, as illustrated in Fig~\ref{fig:workflow}. We begin by selecting a diverse set of prompts, which are then processed by various LLMs to produce soft outputs. These outputs undergo a similarity evaluation using word vector similarity techniques to assess their alignment with the intended sentiment. This systematic approach allows for a comprehensive analysis of the models' performance, offering insights into their ability to capture nuanced sentiment. The workflow's design not only ensures thoroughness in evaluation but also facilitates the identification of specific areas where LLMs excel or require improvement, ultimately contributing to more targeted advancements in sentiment analysis capabilities. Our research not only contributes to the ongoing discourse surrounding LLM evaluation but also highlights the necessary enhancements required to bolster the sentiment sensitivity of these models.

Our contributions are summarized as follows:

\begin{itemize}
\item Incorporating the sentiment analysis, we develop and introduce the ‘Sentiment Knowledge Workflow’ for LLMs. This innovative framework is pivotal in defining and advancing the self-awareness capacities of LLMs.
\item We evaluate the sentiment sensitivity of a diverse array of LLMs across multiple public datasets. Our comprehensive analysis reveals that while these models exhibit a basic ability to detect sentiment, there are significant discrepancies in their accuracy and consistency. These findings underscore the need for further refinements in their training processes to improve their capacity to recognize and respond to subtle emotional cues more effectively.
\end{itemize}

\begin{figure}[t]
    \centering
    \includegraphics[width=1.0\linewidth]{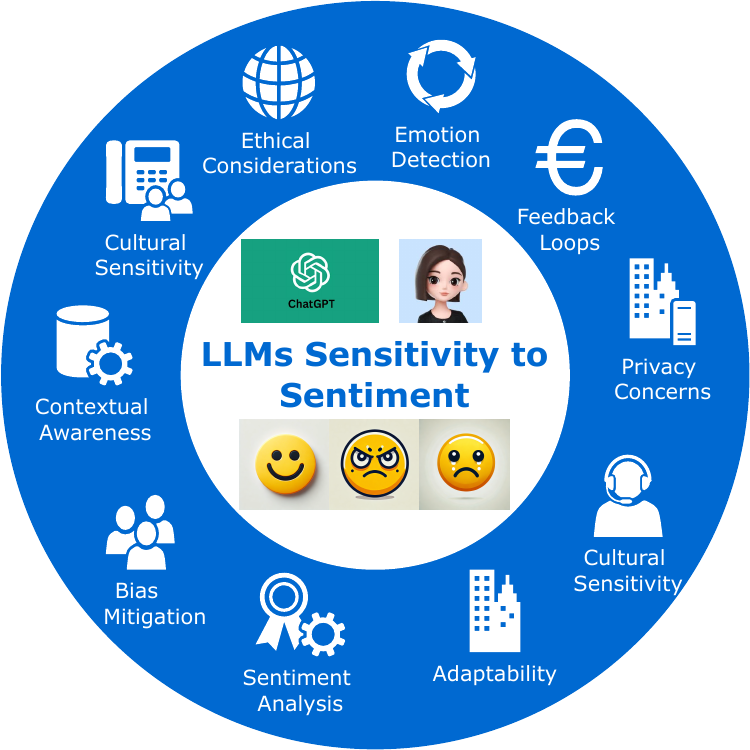}
    \caption{Illustration of the LLMs sensitivity to sentiment topic.}
    \label{fig:sentiment}
\end{figure}


\section{Related Works}\label{section:related}

\paragraph{LLMs Development.} The development of Large Language Models (LLMs) \citep{chen2024entity, zhang2025surveygraphretrievalaugmentedgeneration} represents a significant milestone in the field of artificial intelligence, particularly in natural language processing (NLP) \citep{yuan2023revisiting, yang2024harnessing}. Originating from earlier efforts in neural networks and deep learning, LLMs have evolved rapidly, driven by advancements in computational power, algorithmic innovations, and the availability of vast amounts of textual data. These models, exemplified by architectures like GPT \citep{floridi2020gpt, achiam2023gpt} and BERT \citep{devlin-etal-2019-bert}, are trained on diverse and extensive datasets, enabling them to generate human-like text, perform complex language tasks, and even demonstrate an understanding of context and nuance. The scaling of model parameters \citep{zhang2024scaling} and data has been a crucial factor in enhancing the capabilities of LLMs, allowing them to achieve state-of-the-art performance across a wide range of applications, from translation \citep{lu2024llamax} and summarization \citep{tang2024tofueval} to more specialized tasks like code generation \citep{ugare2024improving, zheng2024towards} and creative writing \citep{gomez2023confederacy}. As research continues, LLMs are poised to further revolutionize how we interact with and understand language in both digital and real-world environments.

\paragraph{LLMs Sentiment Capability.} Large Language Models (LLMs) have increasingly been designed to understand and emulate human emotions \citep{zou2024pilot}, enhancing their role in more nuanced and empathetic communication \citep{sorin2023large, hasan2024llm}. These models are trained on vast datasets that include emotionally rich language, enabling them to recognize and generate text that reflects various emotional tones. By interpreting subtle cues \citep{shukla2023raphael} in language, such as word choice, tone, and context, LLMs can respond in ways that align with the emotional state of the user. This emotional capability is particularly valuable in applications like virtual assistants \citep{vu2024gptvoicetasker}, mental health support \citep{lai2023psy}, and customer service \citep{pandya2023automating}, where understanding and responding to emotions is crucial for effective interaction. However, the development of these capabilities also raises important ethical considerations, as LLMs must navigate complex emotional landscapes without reinforcing biases or generating inappropriate responses. As this technology continues to advance, the emotional intelligence of LLMs \citep{sabour2024emobench, wang2023emotional} is expected to become increasingly sophisticated, allowing for more personalized and empathetic interactions between humans and machines.

\section{Background}

\subsection{Self-awareness.}\label{section:background}

Self-awareness \citep{wang2024mm, yin2023large, liu2024trustworthiness} refers to the ability to recognize and comprehend one's own existence, emotions, thoughts, and behaviors. It encompasses an understanding of one's identity, abilities, strengths, and weaknesses, as well as an awareness of one's role and influence within various social and environmental contexts. Self-awareness \citep{camacho2012self, hall2004self, ortiz2012awareness} can be further categorized into several key aspects, including:

\begin{enumerate}
    \item \emph{Personal Identity Awareness}. Knowing who you are, including your name, age, gender, occupation, interests, and hobbies.
    \item \emph{Sentiment Awareness}. The ability to identify and understand your own emotional states, such as happiness, sadness, anger, and fear.
    \item \emph{Cognitive Self-awareness}. Being aware of and reflecting on your own thoughts and beliefs, including how you make decisions, solve problems, and perceive the world.
    \item \emph{Social Self-awareness}. Understanding your role and status in society, as well as being aware of how others perceive and expect from you.
    \item \emph{Physical Self-awareness}. Recognizing your physical state and sensations, including your appearance, health, and bodily movements.
\end{enumerate}

Self-awareness is a unique characteristic of humans that enables individuals to reflect on their actions, set goals, adjust behavior to adapt to changing environments, and interact effectively in complex social settings. Developing self-awareness can be achieved through self-reflection, psychological counseling, meditation, and communication with others. In this paper, we focus on the sentiment part.

\subsection{Evaluation of LLMs.}

\begin{figure*}[t]
    \centering
    \includegraphics[width=1.0\linewidth]{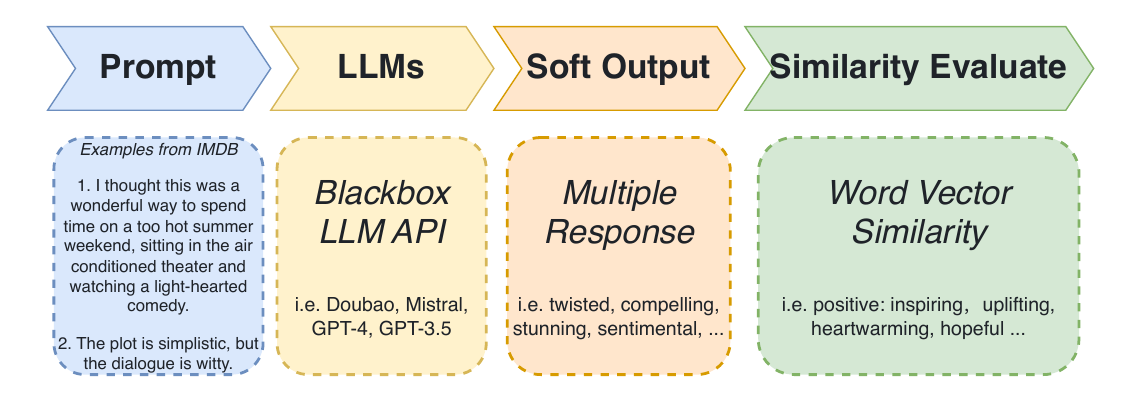}
    \caption{A complete workflow: from prompt engineering to LLM evaluation.}
    \label{fig:workflow}
\end{figure*}

We perform a thorough evaluation of the LLMs using a detailed question-answer workflow, as illustrated in Figure~\ref{fig:workflow}. The process begins with the creation of precise prompts, leading to the generation of initial outputs by the LLMs. These outputs are then analyzed for similarity. The workflow further includes evaluating multiple word vector similarities and categorizing responses based on emotional tones such as stunning, sentimental, positive, inspiring, uplifting, heartwarming, and hopeful. The evaluation concludes with an in-depth assessment of the overall performance and effectiveness of the LLMs.

Various metrics based on multiple-choice questions have been utilized in prominent benchmarks such as CommonsenseQA \citep{talmor2018commonsenseqa}, HellaSwag \citep{zellers2019hellaswag}, and MMLU \citep{hendrycks2020measuring}. These benchmarks have laid the groundwork for evaluating the accuracy of knowledge within language models by focusing on the correct responses to these questions. Building on the methodologies employed in these foundational studies \citep{pan2023llms}, our approach extends their insights by leveraging questions from our specific target tasks. These questions, which are designed to be seamlessly integrated into our evaluation framework, allow us to assess not only the accuracy of the responses but also the depth of understanding and reasoning capabilities exhibited by the models. By incorporating these metrics, we aim to provide a comprehensive evaluation that mirrors the rigor of the original benchmarks while adapting them to the nuanced requirements of our tasks.

\section{Experimental Settings}\label{section:experiment}


\subsection{Dataset.} 

In this study, we utilized three publicly available datasets: Sentiment140, MyPersonality, and IMDB Reviews. The specific details of each dataset are outlined below.

\begin{itemize}
    \item Sentiment140\footnote{\url{https://huggingface.co/datasets/stanfordnlp/sentiment140}}. It is a dataset developed by Stanford University for sentiment analysis research. It consists of 1.6 million tweets collected from Twitter, each labeled as positive, negative, or neutral. What sets Sentiment140 apart is its unique approach to labeling: it uses emoticons in the tweets (such as :-) or :-( ) as sentiment indicators, which are then removed to create a more authentic representation of social media content. This dataset is particularly valuable for handling the challenges of noisy text, including abbreviations, spelling errors, and informal language typical of Twitter. Sentiment140 is widely used for training and evaluating sentiment analysis models, especially those designed to analyze social media data.
    \item Mypersonality\footnote{\url{https://www.psychometrics.cam.ac.uk/productsservices/mypersonality}}. It is a well-known dataset in the fields of psychology and data science, originally created by researchers at the University of Cambridge in 2007. It was generated through an online personality test application hosted on the Facebook platform, where users could take various personality assessments and voluntarily share their Facebook data. This data includes profile information, social network activities, and the results of psychological assessments like the "Big Five Personality Traits" (OCEAN model). Mypersonality offers a unique opportunity for researchers to study the relationship between social media behavior and personality traits. Although the dataset has been controversial due to privacy concerns and data collection methods, it remains a valuable resource for research in psychology and social network analysis.
    \item IMDB Reviews\footnote{\url{https://huggingface.co/datasets/stanfordnlp/imdb}}. It is a widely-used sentiment analysis dataset composed of movie reviews from the Internet Movie Database (IMDb). The dataset typically includes 50,000 reviews, each labeled as either positive or negative, and is used for text classification tasks, particularly sentiment analysis. Unlike shorter text datasets, IMDB Reviews features longer reviews, rich with semantic information such as opinions, emotions, and arguments. This makes it an ideal dataset for evaluating and training deep learning models that need to handle more complex semantics and contextual information. Given IMDb’s global reach, the dataset encompasses a wide range of expressions and cultural backgrounds, making it valuable for testing the generalization capabilities of sentiment analysis models.
\end{itemize}

\subsection{Tasks} 

The focus of this paper is on the task of multi-label classification, where each input query \( q \) can be linked to multiple labels simultaneously, rather than being limited to a single label. Formally, given a query \( q \), the model outputs a set of labels \( y = \{y_1, y_2, \dots, y_k\} \), with each \( y_i \in \{0, 1\} \) representing the presence (1) or absence (0) of the corresponding label. Large language models (LLMs) address this task by leveraging their advanced ability to comprehend complex textual contexts, enabling them to effectively predict multiple relevant labels by identifying and capturing intricate patterns embedded within the data.

\begin{algorithm*}
\small
\caption{Evaluation Process of LLMs' Sentiment Analysis}
\textbf{Input:} Questions (Prompt) $\mathcal{Q}$, candidate options set $\mathcal{O} = \{ o_1, o_2, \ldots, o_n \}$, pretrained embedding model $\mathcal{M}$, LLM API $\mathcal{K}$. \\
\textbf{Output:} Option $\hat{o} \in \mathcal{O}$ with highest probability (LLM response).
\begin{algorithmic}[1]
\STATE clean prompt $\mathcal{Q}^{\prime}$ from $\mathcal{Q}$
\STATE compute embedding of candidate labels $\mathcal{E} = \{e_i\}_{i=1}^n = \{ \mathcal{M}(o) \mid o \in \mathcal{O} \}$
\STATE \textcolor{red}{[Rule1]} obtain one response from LLM API $\mathcal{R} = \{r\} = \mathcal{K}(\mathcal{Q}^{\prime})$
\STATE \textcolor{red}{[Rule2]} obtain multiple soft response from LLM API $\mathcal{R} = \{r_1, \ldots, r_m\} = \mathcal{K}(\mathcal{Q}^{\prime})$
\STATE compute embedding $e_r = \mathcal{M}(\mathcal{R})$
\STATE compute similarity score s = [$\operatorname{SIMILARITY}(e_r, e) \mid e \in \mathcal{E}$]
\STATE \textcolor{red}{[Optional]} probs = $\operatorname{SOFTMAX}(s)$ \hfill $\triangleright$ normalize the similarity score
\STATE candidate\_answer = dict([(i, op) for i, op in enumerate($\mathcal{O}$)])
\STATE answer = candidate\_answer[np.argmax(probs)] \hfill $\triangleright$ choose the highest similarity score
\end{algorithmic}
\end{algorithm*}

\begin{table*}[t]
\small
\centering
\caption{Evaluation Results of LLMs on Sentiment140 dataset. 
}
\label{table:140}
\begin{tabular}{lccccc}
\toprule
\textbf{Models} & \textbf{Accuracy} & \textbf{Precision} & \textbf{Recall} & \textbf{F1-score} & \textbf{ROC-AUC} \\
\midrule
\midrule
GPT-3.5-turbo & 0.62 & 0.85 & 0.68 & 0.78 & 0.89\\
GPT-4 & 0.64 & 0.67 & 0.68 & 0.73 & 0.81\\
GPT-4o & 0.72 & 0.72 & 0.59 & 0.65 & 0.78\\
llama\_7b\_v2 & 0.50 & 0.66 & 0.61 & 0.59 & 0.62 \\
llama\_8b\_v3 & 0.72 & 0.87 & 0.71 & 0.76 & 0.87 \\
Mistral\_7b & 0.71 & 0.88 & 0.69 & 0.76 & 0.88\\
Doubao & 0.75 & 0.75 & 0.56 & 0.66 & 0.75\\
Doubao-pro & 0.77 & 0.76 & 0.56 & 0.69 & 0.82\\
\bottomrule
\end{tabular}
\end{table*}

\begin{table*}[t]
\small
\centering
\caption{Evaluation Results of LLMs on Mypersonality dataset. 
}
\label{table:Mypersonality}
\begin{tabular}{lccccc}
\toprule
\textbf{Models} & \textbf{Accuracy} & \textbf{Precision} & \textbf{Recall} & \textbf{F1-score} & \textbf{ROC-AUC} \\
\midrule
\midrule
GPT-3.5-turbo & 0.52 & 0.55 & 0.80 & 0.61 & 0.77 \\
GPT-4 & 0.82 & 0.53 & 0.76 & 0.65 & 0.79 \\
GPT-4o & 0.80 & 0.69 & 0.75 & 0.65 & 0.81 \\
llama\_7b\_v2 & 0.38 & 0.39 & 0.52 & 0.38 & 0.39 \\
llama\_8b\_v3 & 0.57 & 0.48 & 0.86 & 0.61 & 0.77 \\
Mistral\_7b & 0.59 & 0.45 & 0.83 & 0.59 & 0.80 \\
Doubao & 0.72 & 0.66 & 0.79 & 0.64 & 0.72 \\
Doubao-pro & 0.75 & 0.70 & 0.75 & 0.66 & 0.75 \\
\bottomrule
\end{tabular}
\end{table*}

\begin{table*}[t]
\small
\centering
\caption{Evaluation Results of LLMs on IMDB dataset. 
}
\label{table:imdb}
\begin{tabular}{clccccc}
\toprule
& \textbf{Models} & \textbf{Accuracy} & \textbf{Precision} & \textbf{Recall} & \textbf{F1-score} & \textbf{ROC-AUC} \\
\midrule
\midrule
\multirow{8}{*}{\textbf{rare input}}
& GPT-3.5-turbo & 0.58 & 0.45 & 0.59 & 0.53 & 0.57\\
& GPT-4 & 0.65 & 0.41 & 0.56 & 0.49 & 0.60\\
& GPT-4o & 0.62 & 0.47 & 0.61 & 0.49 & 0.63\\
& llama\_7b\_v2 & 0.50 & 0.24 & 0.50 & 0.31 & 0.50 \\
& llama\_8b\_v3 & 0.67 & 0.49 & 0.66 & 0.42 & 0.65 \\
& Mistral\_7b & 0.60 & 0.44 & 0.60 & 0.42 & 0.60 \\
& Doubao & 0.71 & 0.55 & 0.70 & 0.54 & 0.70\\
& Doubao-pro & 0.72 & 0.55 & 0.72 & 0.54 & 0.70\\
\midrule
\multirow{8}{*}{\textbf{processed input}}
& GPT-3.5-turbo & 0.63 & 0.42 & 0.64 & 0.54 & 0.52\\
& GPT-4 & 0.67 & 0.43 & 0.58 & 0.45 & 0.59 \\
& GPT-4o & 0.60 & 0.50 & 0.61 & 0.52 & 0.68 \\
& llama\_7b\_v2 & 0.51 & 0.26 & 0.50 & 0.34 & 0.50 \\
& llama\_8b\_v3 & 0.53 & 0.27 & 0.50 & 0.38 & 0.50 \\
& Mistral\_7b & 0.61 & 0.46 & 0.60 & 0.46 & 0.61 \\
& Doubao & 0.72 & 0.55 & 0.70 & 0.55 & 0.68\\
& Doubao-pro & 0.74 & 0.55 & 0.71 & 0.57 & 0.70\\ 
\bottomrule
\end{tabular}
\end{table*}

\subsection{Evaluation.} 

Metrics for multi-class classification are used to evaluate the performance of models that predict multiple classes. Key metrics include accuracy, precision, recall, and F1 score, each providing insights into different aspects of the model's performance. Accuracy measures the proportion of correctly predicted samples out of the total samples. Precision and recall assess the model's performance for each class, with precision indicating how many of the predicted positives are actually correct, and recall showing how many actual positives were correctly identified. The F1 score, the harmonic mean of precision and recall, offers a balanced evaluation of the model's effectiveness.

\subsection{Baselines.} 

We utilize well-established LLMs, such as LLaMA, as baseline models in our experiments. Unless specified otherwise, all baseline models are implemented using the parameters provided in their respective original APIs.

\begin{itemize}
    \item ChatGPT\footnote{\url{https://openai.com/}}. It is an advanced conversational AI developed by OpenAI, designed to generate human-like text based on given prompts. It comes in various versions, including GPT-3.5-turbo, GPT-4, and GPT-4-turbo. GPT-3.5-turbo offers efficient performance and responsiveness, making it well-suited for a wide range of applications. GPT-4, a more powerful and sophisticated model, provides enhanced language understanding and generation capabilities, ideal for complex tasks. GPT-4-turbo further optimizes performance, delivering faster responses while maintaining the high quality and depth of GPT-4's output.
    \item LLaMA\footnote{\url{https://huggingface.co/docs/transformers/main/model_doc/llama3}} (Large Language Model Meta AI). It is a family of advanced language models developed by Meta, designed to generate and understand human-like text. LLaMA models are known for their efficiency and effectiveness in various natural language processing tasks. Within this family, Mistral\footnote{\url{https://huggingface.co/mistralai/Mistral-7B-v0.1}} is a notable variant that focuses on optimizing performance and resource usage, offering high-quality outputs while being more computationally efficient. Mistral represents a key innovation within the LLaMA series, combining state-of-the-art language generation with enhanced scalability and accessibility for diverse applications.
    \item Doubao\footnote{\url{https://huggingface.co/doubao-llm}}. It is an advanced language model designed to excel in a wide range of natural language processing tasks, from text generation and sentiment analysis to machine translation. Built with cutting-edge deep learning techniques and trained on extensive data, Doubao captures intricate linguistic patterns and contextual meanings, enabling it to generate human-like text across various contexts. Its robust performance and versatility make it a valuable tool for industries such as customer service, content creation, academic research, and data-driven decision-making. Doubao's capabilities contribute significantly to the advancement of AI technologies and our understanding of language.
\end{itemize}

\section{Results and Insights}\label{section:discussions}

\subsection{Results}

We present a comprehensive overview of our evaluation results across three different datasets, which are shown in Table~\ref{table:140}, Table~\ref{table:Mypersonality}, and Table~\ref{table:imdb}. These tables encompass a range of models and configurations, offering detailed insights into their performance. Additionally, we provide a demonstration analysis in Table~\ref{table:prompt}, where we apply various prompt templates to the IMDB dataset. For further clarity, the specific details of the prompt templates can be found in Table~\ref{table:prompt_id}. In the following sections, we discuss key findings and insights, addressing each observation individually for a more in-depth understanding.

\subsection{Interesting Insights}

\paragraph{Insight 1.} Some LLMs possess a unique ability to be sensitive to sentiment. (Refer to Table \ref{table:140}, \ref{table:Mypersonality} and \ref{table:imdb})

\paragraph{Insight 2.} Processing prompts cannot obscure or eliminate LLMs ability to detect sentiment with neutral prompts. (Refer to Table \ref{table:imdb} and \ref{table:prompt})

\paragraph{Insight 3.} Different versions of the same LLM can exhibit varying behaviors and performance. (Refer to Table \ref{table:140}, \ref{table:Mypersonality} and \ref{table:imdb})

\begin{table}[h]
\small
\centering
\caption{Evaluation Results of Doubao on IMDB dataset with various prompt template. 
}
\label{table:prompt}
\begin{tabular}{llccc}
\toprule
\textbf{Models} & \textbf{PROMPT ID} & \textbf{P} & \textbf{R} & \textbf{F1} \\
\midrule
Doubao & PROMPT\_NEUTRAL & 0.55 & 0.70 & 0.55 \\
Doubao & PROMPT\_POSITIVE & 0.25 & 0.50 & 0.33 \\
Doubao & PROMPT\_NEGATIVE & 0.26 & 0.50 & 0.34 \\
\midrule
Doubao-pro & PROMPT\_NEUTRAL & 0.55 & 0.71 & 0.57\\
Doubao-pro & PROMPT\_POSITIVE & 0.24 & 0.50 & 0.34 \\
Doubao-pro & PROMPT\_NEGATIVE & 0.25 & 0.50 & 0.32 \\
\bottomrule
\end{tabular}
\end{table}

\section{Discussion}

The evaluation results across different datasets and models shed light on the varying capabilities of LLMs, particularly when it comes to sentiment detection.

LLMs exhibit a certain sensitivity towards sentiment that appears to be a unique characteristic across multiple models, as shown in Table \ref{table:140}, Table \ref{table:Mypersonality}, and Table \ref{table:imdb}. For example, the Doubao-pro model consistently performs well in sentiment tasks, demonstrating strong scores in both precision and recall across multiple datasets. This suggests that the underlying architecture of some LLMs might be more adept at capturing emotional subtleties in text, despite the varying nature of the input data. This sensitivity indicates that certain LLMs can be fine-tuned or selected specifically for sentiment-related tasks, even in a competitive landscape of multiple LLM options.

The ability of LLMs to detect sentiment is not easily obscured by prompt processing, particularly when dealing with neutral prompts, as evident from Table \ref{table:prompt}. Doubao-pro maintains a relatively high performance with neutral prompts, despite changes in input structure. This suggests that the model's internal mechanisms for identifying sentiment are robust enough to operate even when the prompt is neutral, implying that sentiment detection in LLMs may be deeply ingrained in the model's learned representations, rather than being highly sensitive to prompt formulations. This highlights the model's flexibility and adaptability in different real-world scenarios where the exact phrasing of the input may vary.

The comparison between different versions of the same LLM, such as the Doubao and Doubao-pro models, reveals that even slight modifications to the architecture or training procedures can lead to notable differences in performance, as shown in Table \ref{table:140}, Table \ref{table:Mypersonality}, and Table \ref{table:imdb}. Doubao-pro consistently outperforms its predecessor across multiple datasets and metrics, showing that model refinement plays a crucial role in enhancing the ability of LLMs to perform on sentiment tasks. This variability underscores the importance of continuous model development and experimentation to achieve optimal results in practical applications. These detailed insights together provide a deeper understanding of how LLMs behave under various conditions and prompt configurations, suggesting potential strategies for optimizing LLMs for sentiment analysis tasks in diverse applications.

\bibliography{custom}

\onecolumn
\appendix

\section*{Appendix}
\section{Prompt Design}
\label{sec:Prompt}

For each task, we apply a consistent prompt engineering template to generate the input prompt. The templates are listed below.

\begin{table}[h]
\small
\centering
\caption{Prompt Summarizing.
}
\label{table:prompt_id}
\begin{tabular}{llp{9cm}}
\toprule
\textbf{Dataset} & \textbf{Prompt ID} & \textbf{Prompt Content} \\
\midrule
Sentiment140 & PROMPT\_NEUTRAL 
& \texttt{You are a psychologist. Please analysis the general sentiment of the tweet based on the following text. You should also provide a brief description of the tweet with some more words or sentences. 
\newline\newline \{TWEET CONTENT\} } \\
\midrule
Mypersonality & PROMPT\_NEUTRAL
& \texttt{You are a educational psychologist. Please analysis the personality of the person based on the following detailed information. You should also provide a brief description of the person's personality with some more words or sentences. 
\newline\newline\{PERSON DETAILED INFORMATION\} } \\
\midrule
& PROMPT\_NEUTRAL 
& \texttt{You are a film critic. Please classify the general sentiment of the film based on the following reviews as either "positive" or "negative." You should also provide a brief description of the movie with some more words or sentences, such as "positive, this movie looks wonderful!" or "negative, this movie sucks." 
\newline\newline \{REVIEW CONTENT\} } \\
IMDB & PROMPT\_POSITIVE & \texttt{You are a film critic with more positive attitude. Please classify the general sentiment of the film based on the following reviews as either "positive" or "negative." You should also provide a brief description of the movie with some more words or sentences, such as "positive, this movie looks wonderful!" or "negative, this movie sucks." 
\newline\newline \{REVIEW CONTENT\} } \\
& PROMPT\_NEGATIVE & \texttt{You are a film critic with more negative attitude. Please classify the general sentiment of the film based on the following reviews as either "positive" or "negative." You should also provide a brief description of the movie with some more words or sentences, such as "positive, this movie looks wonderful!" or "negative, this movie sucks." 
\newline\newline \{REVIEW CONTENT\} } \\
\bottomrule
\end{tabular}
\end{table}

\end{document}